\documentclass[runningheads]{llncs}
\usepackage{graphicx}
\usepackage{color}

\usepackage{hyperref}       
 
\def\figureautorefname{Fig.}

\usepackage{amsmath}
\usepackage{latexsym}
\usepackage{xspace}
\usepackage{amssymb}
\usepackage{blindtext}
\usepackage[colorinlistoftodos]{todonotes}
\usepackage{color}

\usepackage{comment}

\newcommand{\dfr}{\ensuremath{\mapsto}}
\newcommand{\efr}{\ensuremath{\leadsto}}

\newcommand{\pot}{partially ordered trace\xspace}

\begin{document}
\title{Interactive Multi-Interest Process Pattern Discovery}
\author{Mozhgan Vazifehdoostirani\inst{1} \and
Laura Genga \inst{1} \and
Xixi Lu \inst{2}\and 
Rob Verhoeven\inst{3, 4, 5} \and
Hanneke van Laarhoven\inst{4, 5} \and
Remco Dijkman\inst{1}}
\institute{Eindhoven University of Technology, the Netherlands \and
Utrecht University, The Netherlands \and
Netherlands Comprehensive Cancer Organisation (IKNL), 
the Netherlands \and
Amsterdam UMC location University of Amsterdam, 
the Netherlands \and
Cancer Center Amsterdam, Cancer Treatment and Quality of Life, the Netherlands
}
\authorrunning{M. Vazifehdoostirani et al.}
\maketitle              
%
\begin{abstract} 
Process pattern discovery methods (PPDMs) aim at
identifying patterns of interest to users. Existing PPDMs typically are unsupervised and focus on a single dimension of interest, such as discovering frequent patterns.
We present an interactive multi-interest-driven framework for process pattern discovery aimed at identifying patterns that are optimal according to a multi-dimensional analysis goal. The proposed approach is iterative and interactive, thus taking experts' knowledge into account during the discovery process.
The paper focuses on a concrete analysis goal, i.e., deriving process patterns that affect the process outcome. We evaluate the approach on real-world event logs in both interactive and fully automated settings. The approach extracted meaningful patterns validated by expert knowledge in the interactive setting. Patterns extracted in the automated settings consistently led to prediction performance comparable to or better than patterns derived considering single-interest dimensions without requiring user-defined thresholds.


\end{abstract}
\section{Introduction}

Process pattern discovery methods (PPDMs) 
aim to discover process patterns that are \textit{of interest} for the human analyst, where a process pattern corresponds to a set of process activities (possibly annotated with additional data) with their ordering relations. The interest of a pattern is usually computed according to one or more functions. Previous studies highlighted how these techniques often uncovered interesting behaviors that would otherwise remain hidden in start-to-end process models~\cite{tax2016mining}. 

Several approaches have been proposed to discover process patterns 
from a given event log~\cite{tax2016mining,bose2009trace,gunther2007fuzzy}
and employed them in several applications, for instance, event abstraction \cite{mannhardt2017unsupervised}, or trace classification \cite{vazifehdoostirani2022encoding}. However, most of these approaches focus on a single interest dimension. In particular, they usually aim to detect \textit{frequent} patterns, which often leads to
the generation of a multitude of non-interesting patterns and possibly the missing of interesting but infrequent ones \cite{tax2018interest}. As pointed out by recent studies in the pattern mining field~\cite{fang2020mining}, the concept of \textit{interest} of a pattern is often linked to multiple dimensions, some of which may be in conflict with each other. These considerations also hold in the process domain since processes emerge from the interplay of multiple factors, highlighting the need for multi-dimensional thinking in process analysis~\cite{fahland2022multi}.
Few PPDMs introduced a broader notion of interest, 
either by allowing the user to define cut-off thresholds for several metrics~\cite{tax2016mining}, which are then aggregated to rank the obtained set of patterns, or by directly using a composite metric during the pattern generation phase~\cite{tax2018interest,diamantini2016behavioral}.
However, these solutions offer limited support in dealing with a multi-dimensional notion of pattern interest. Defining appropriate cut-off thresholds for different and conflicting metrics is a non-trivial decision that strongly impacts the obtained results. Furthermore, aggregating multiple dimensions in a single one leads to a single ranked collection of patterns which depends on the aggregation setting and hides the interplay of the different dimensions.
To deal with this complexity, the detection of process patterns should be expressed as a \textit{multi-objective} problem.

Beside the \textit{multi-objective} challenge, most PPDMs are unsupervised and suffer from pattern explosion in real-life event logs. Previous studies showed that 
leveraging expert domain knowledge can avoid or mitigate the pattern explosion issue~\cite{lu2017semi,atzmueller2019framework}. A semi-supervised PPDM~\cite{lu2017semi} was proposed for users to manually select and extend patterns.
However, the approach still relies exclusively on frequency-based metrics. 
Also, the burden of the selection and extension of discovered patterns is left to the user as a manual task without much guidance.

In this work, we introduce the IMPresseD framework (\textbf{I}nteractive \textbf{M}ulti-interest \textbf{Pr}oc\textbf{ess} Pattern \textbf{D}iscovery) for process pattern discovery. IMPresseD is designed to derive interesting and easily interpretable patterns for the end users by combining different strategies. First,
the framework allows users to define different \emph{interest functions} to measure the interest of patterns, supporting customizable multi-dimensional analysis goals. In this way, the user has more control over the measures of relevance that they use, which is expected to lead to patterns that are indeed considered meaningful by end users. Multi-optimization strategies are used to allow the user to go over far fewer patterns than the ones obtained by threshold-dependent strategies to identify the relevant ones. 
The framework supports an in-depth analysis of the pattern characteristics, which also considers the characteristics of the process executions in which the pattern occurs. Finally, the approach is iterative and interactive. At each step, the user is presented with the process patterns that are best according to the user-defined interest functions, and they can select the ones to expand further. 

To showcase the framework's usefulness, we also discuss how to use it with a concrete analysis goal, i.e., \textit{deriving process patterns affecting the process outcome}. This is inherently a complex problem for which different aspects need to be considered.
Furthermore, to the best of our knowledge, most outcome-oriented pattern detection approaches do not support a multi-dimensional analysis. Given this concrete analysis goal, we carried out a two-fold evaluation to validate our approach. First, we use a real-world case study in healthcare to show the capability of the proposed framework in supporting domain experts in extracting meaningful patterns in an interactive setting. Then, we evaluate our approach in a quantitative experiment to assess the predictive power of the automatically discovered patterns.
We compare the results of our approach with the ones obtained by using a single metric and using the entire pattern set without filtering. The obtained results show that the discovered set of patterns consistently ranked within the top positions, while patterns mined by adopting single metrics led to a more unstable performance. Furthermore, the proposed framework returned a set of patterns significantly smaller than the entire pattern set while preserving a comparable predictive power.

Summing up, the paper contributes to the literature by introducing:
\begin{itemize}
    \item a multi-interest and interactive process pattern discovery framework;
    \item tailored interest functions for discovering process patterns affecting the outcome of the process.
\end{itemize}

The remainder of this paper is organized as follows. Section \ref{I} reviews the relevant related work. Section \ref{II} provides basic concepts used throughout the paper. Section \ref{III} introduces the proposed framework, together with a concrete instantiation of the interest function to support outcome-oriented pattern discovery. Section \ref{IV} presents and discusses 
the evaluation. 
Finally, Section \ref{V} draws the conclusion and delineates some ideas for future studies. 

\section{Related work}\label{I}



Most previous PPDMs take an event log as input and generate patterns based on user-defined thresholds on a set of predefined measures of interest. These approaches vary depending on the type of patterns they aim to extract. Early work focused on discovering sequences of event traces, such as identifying sequences that fit predefined templates~\cite{bose2009abstractions} or using a sequence pattern mining algorithm~\cite{huang2012mining}. More recent research has focused on patterns representing more complex control-flow relationships, for instance,
episodes representing eventually-flow relations~\cite{leemans2014discovery}, or graphs
representing both sequential and concurrent behaviors~\cite{hwang2004discovery,diamantini2016behavioral}. Patterns that represent a more comprehensive set of control-flow relationships, including sequences, concurrency, and choice, are considered in the approach proposed by Tax et al.~\cite{tax2016mining}. This approach has been extended to allow the extraction of patterns 
based on a more general set of utility functions~\cite{tax2018interest}. Taking into account the context in which patterns are observed, Acheli et al. extended previous work to discover contextual behavioral patterns, allowing for insights into the aspects that influence a process conduction~\cite{acheli2021discovering}.

Although these unsupervised PPDMs can uncover interesting patterns, 
they offer little or no support for multi-dimensional analysis goals involving possibly conflicting dimensions.
Furthermore, they do not incorporate user knowledge, which often results in the return of uninteresting patterns. A possible mitigation strategy to this problem consists in keeping \lq \lq humans in the loop", as observed by previous authors. For instance, Benevento et al. showed potential improvements in the quality and clarity of the process models by employing interactive process discovery in modeling healthcare processes compared to traditional automated discovery techniques~\cite{benevento2019evaluating,benevento2022can}. 
Within the PPDMs domain, a semi-supervised approach is proposed for discovering process patterns which involves the user in the pattern extraction process~\cite{lu2017semi}. However, this approach only exploits frequency-based interest functions based on user-defined thresholds.
\section{Preliminaries}\label{II}
In this section, we recall the basic concepts needed to introduce our framework.

\begin{definition}[Event]
Let $\mathcal{AC}$ be the universe of activities, $\mathcal{C}$ be the universe of case identifiers, $\mathcal{T}$ be the time domain, 
and $\mathcal{D}_1, \mathcal{D}_2, ..., \mathcal{D}_m $ be the sets of additional attributes with $i\;{\in}\;[1,m]$, $m\;\in\;\mathbb{Z}$.
An event is a tuple of $e=(a,c,t,d_1,\dots,d_{m})$, where $a\;{\in}\;\mathcal{AC}$, $c\;{\in}\;\mathcal{C}$, $t\;{\in}\;\mathcal{T}$ and
$d_{i}\;{\in}\;\mathcal{D}_i$.
\end{definition}

\begin{definition}[Trace, event log]\label{def:log}
A \emph{trace} $\sigma = \langle e_1, \cdots, e_n\rangle$ is a finite non-empty sequence of events $e_1, \cdots, e_n$ in which their timestamp does not decrease. 
Let $\mathcal{S}$ denote the universe of all possible traces, an \emph{event log} can be defined as $L = \{\sigma_1, \sigma_2, \cdots, \sigma_n\}$ $\subseteq S$ which is a set of traces.
\end{definition}

We use $E_{\sigma}$ for the set of events in trace $\sigma$.
We define $\pi_{act}(e)$, $\pi_{time}(e)$, $\pi_{case}(e)$, and $\pi_{d_i}(e)$ to return the activity, timestamp, case identifier and the attribute $d_i$
associated with $e$, respectively.


A well-known issue of log traces is that they flatten the real ordering relations among process events, hiding possible concurrency~\cite{leemans2023partial}. Since we intend to discover patterns representing both sequential and concurrent relations, we convert log traces
in so-called \textit{partially ordered traces}.
It is possible to derive partially ordered traces from fully ordered traces by using a conversion oracle function obtained from expert knowledge or data analysis~\cite{DIAMANTINI2016101,lu2015conformance}.


\begin{definition}[Partially ordered trace]\label{def:pot}
Given a conversion oracle function $\varphi$ and a log trace $\sigma$, a \emph{\pot} $\varphi(\sigma) = (E_{\sigma}, \prec_{\sigma})$ is a Directed Acyclic Graph (DAG), where $E_{\sigma}$ and $\prec_{\sigma}\; \in E_{\sigma} \times E_{\sigma}$ corresponds to the set of nodes and edges, respectively. We define matrix $A_{\varphi(\sigma)}$ as an upper triangular adjacency matrix that specifies directed edges from $e$ to $e'$, with $e, e'\in E_{\sigma}$. Also,
$R_{\varphi(\sigma)}$ is the reachability matrix derived from $A_{\varphi(\sigma)}$ to represent all possible paths from $e$ to $e'$ of length $l$ such that $2\leq l \leq \mid \sigma \mid-1$ .
For each pair of events $e,e^{'} \in E_{\sigma}$, such that $e \neq e'$, we define the following ordering relations:
\begin{itemize}
    \item if $R_{\varphi(\sigma)}(e, e') \neq 0$, $e'$ \emph{eventually follows} $e'$,
    \item if $A_{\varphi(\sigma)}(e, e') \neq 0$, $e'$ \emph{directly follows} $e$,
    \item if $R_{\varphi(\sigma)}(e, e') = 0$ and $R_{\varphi(\sigma)}(e', e) = 0$, then $e$ is \emph{concurrent} with $e'$.
\end{itemize}

\end{definition}

\begin{definition}[Process pattern]\label{def:pattern}
A process pattern $P = (N, \dfr, \alpha, \beta)$ is a DAG, where:
\begin{itemize}
\item 
$N$ is a set of nodes,
\item 
$\dfr$ is set of edges over $N$ 
\item 
$\alpha$ is a function that assigns a label $\alpha(n)$ to any node $n \in N$,  
\item 
$\beta$ is a foundational pattern for $P$, which means pattern $P$ is extended from pattern $\beta$. 

\end{itemize}

Also, we denote that if $\mid N\mid = 1$, then $\beta$ is \texttt{NULL}, i.e., a single node is considered a pattern without any foundational pattern. 

\end{definition}
Examples of process patterns can be found in
\figureautorefname~\ref{fig:6}. $P_{\theta}^{1}, ..., P_{\theta}^{5}$
all share the same foundational pattern $\theta$, represented by the single node \textit{b}. In turn, pattern $\omega$ is the foundational pattern for $P_{\omega}^{1}$ and $P_{\omega}^{2}$ in the middle column.
Given a process pattern, an \textit{instance} of the pattern is an occurrence of the pattern in a log trace.

\begin{definition}[Pattern instances set]
Let $P = (N, \dfr, \alpha, \beta)$ be a pattern, $\varphi(\sigma) = (E_{\sigma}, \prec_{\sigma})$ a partially ordered trace, $A_{\dfr}$ be an upper triangular adjacency matrix over $N$, and $R_{\efr}$ be the reachability matrix of size {\small $\mid N\mid -1$} derived from $A_{\dfr}$. Given a subset $E' \subseteq E_\sigma$ of nodes in $\varphi(\sigma)$, such that there is a bijective function $I: E' \rightarrow N$, then we define the \emph{pattern instances} of $P$ in $\varphi(\sigma)$ as $PI(P, \varphi(\sigma)) = \{ E' \mid \forall e, e' \in E', A_{\varphi(\sigma)}(e, e') = A_{\dfr}(I(e), I(e')) \wedge  R_{\varphi(\sigma)}(e, e') = R_{\efr}(I(e), I(e')) \wedge \pi_{act}(e) = \alpha(I(e)) \}$. 
The pattern instances set of pattern $P$ over event log $L$ is defined as {\small $PIS(P, L, \varphi) = \bigcup_{\sigma \in L} PI(P, \varphi(\sigma))$}.

\end{definition}

\section{IMPresseD framework}\label{III}
\begin{figure}
    \centering
    \includegraphics[width=\textwidth]{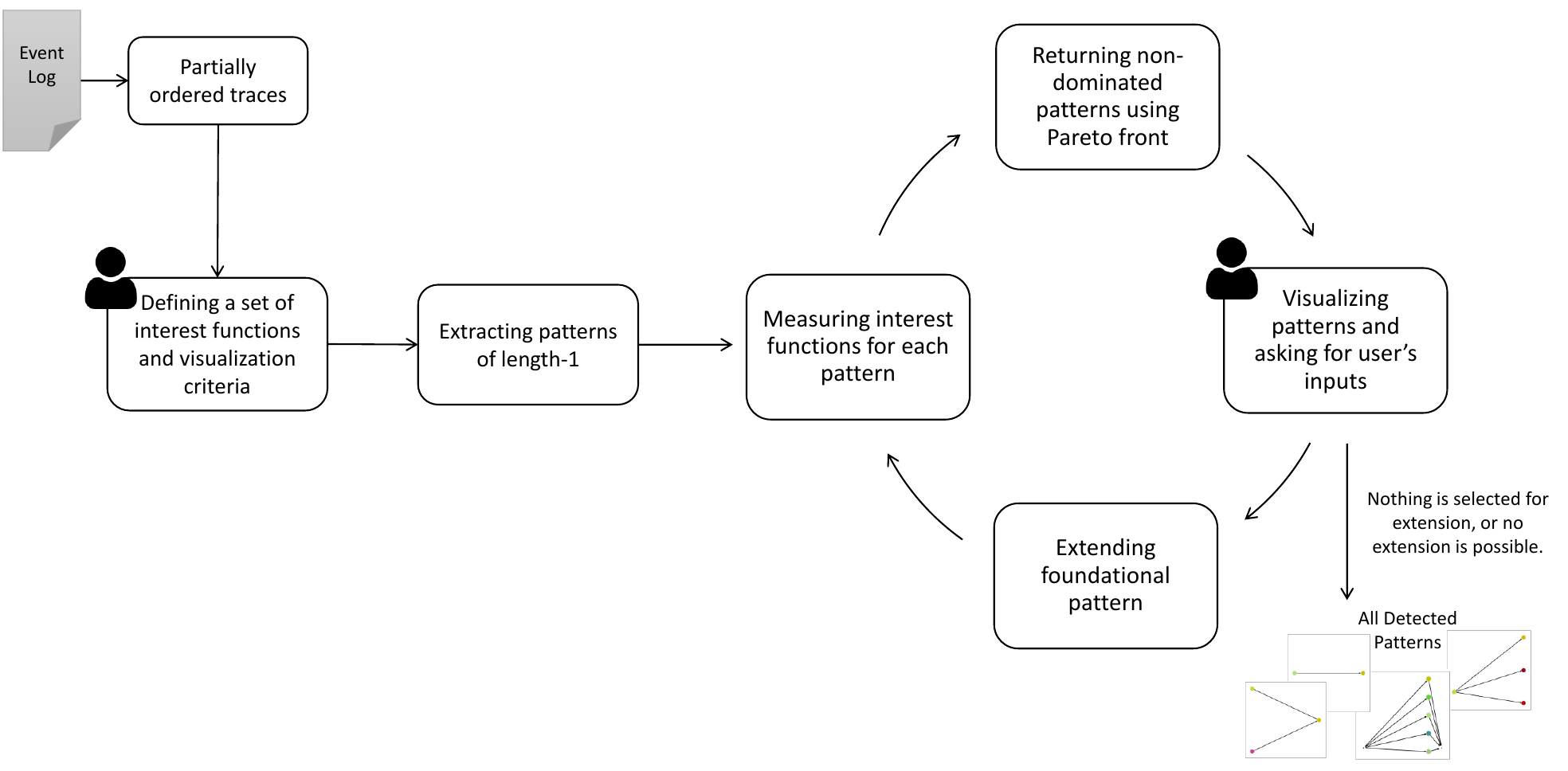}
    \caption{Overview of the IMPresseD framework}
    \label{fig:1}
\end{figure}

Given an event log, the objective of the IMPresseD framework (\figureautorefname~\ref{fig:1}) is to discover the set of process patterns that are best according to multiple interest functions defined by the user. The framework includes the following steps.


\begin{itemize}
    \item[Step 1] Converting all traces in the event log into partially ordered traces using a conversion oracle derived from expert knowledge or data analysis.
    \item[Step 2] Defining the interest functions which fit the users' notion of pattern interestingness based on their analysis goal. Analytical dashboards to visualize the discovered patterns and the computed interest functions are also defined.
    \item[Step 3] Extracting patterns of length-1, i.e., individual activities.
    \item[Step 4] Measuring the interestingness of each discovered pattern through the set of  interest functions defined at \textit{Step 2}.
    \item[Step 5] Returning the set of patterns that are the best according to the interest functions (i.e., non-dominated patterns in the Pareto front).
    \item[Step 6] 
    If the user is satisfied with the current set of patterns or there is no extension possible, the procedure ends. Otherwise, the user selects pattern(s) to extend (i.e., the foundational patterns), and the procedure goes to \textit{Step 7}. 
    \item [Step 7]
    Building all extensions of the foundational patterns and going to \textit{Step 4}.
\end{itemize}

In the remainder of this section, we delve into 
the pattern \textit{selection} (Steps 4 and 5) and \textit{extension} (Step 7). Finally, we show an instantiation of the interest functions (Step 2) using an
analysis goal for the discovery of process patterns affecting the process outcome.



\subsection{Pattern selection}\label{III_2}

Let $\mathcal{P}_{i} = \{P^{1}, P^{2},..., P^{k} \}$ be the set of all patterns discovered 
in $i^{th}$ iteration of the method and let $\mathcal{I} = \{ \mathcal{I}_{1}, \mathcal{I}_{2},..., \mathcal{I}_{m}\}$ be the set of interest functions, where $\forall \mathcal{I}_{k} \in \mathcal{I}$,  $\mathcal{I}_{k}: \mathcal{P}_{i} \rightarrow \mathbb{R}$.
The pattern selection module aims to return the set of patterns ($ P^{*} \subseteq \mathcal{P}_{i}$) that optimize the pre-defined interest functions.
This corresponds to solving a \textit{multi-objective optimization problem (MOP)}.

\begin{sloppypar}
Several approaches have been proposed in the literature to solve a MOP. Note, however, 
that a feasible solution optimizing all objective functions simultaneously usually does not exist. Therefore, the goal is to find the so-called \emph{Pareto Front}, which involves a set of patterns that are not dominated by any other pattern in terms of the multiple interest functions. Informally, solutions on the Pareto front are such that no objective can be improved without worsening at least one of the other objectives. 
In this paper, we use the algorithm proposed by \cite{914855} to filter out dominated patterns.
For any pair of patterns $P^{l}$, $P^{j} \in \mathcal{P}_{i}$, we say that $P^{l}$ dominates $P^{j}$ if and only if: a) $\forall \mathcal{I}_{k} \in \mathcal{I}, \mathcal{I}_{k}(P^{l}) \text{ is no worse than } \mathcal{I}_{k}(P^{j})$; b) $\exists \mathcal{I}_{k} \in \mathcal{I}, \mathcal{I}_{k}(P^{l}) \text{ is strictly better than }\mathcal{I}_{k}(P^{j})$.

\end{sloppypar}
    
   




\subsection{Pattern extension}\label{III_2}
Informally, extending a pattern $P$ means generating a new pattern $P'$ by adding new nodes and edges to $P$ according to a set of \textit{extension rules} applied on partially ordered traces involving at least one instance of the pattern. Formally,
let $\varphi(\sigma) = (E_{\sigma}, \prec_{\sigma})$ be a partially ordered trace, and $P = (N, \dfr, \alpha, \beta)$ be a pattern, in a way that $ \mid PI(P,\varphi(\sigma)) \mid > 0$ and $E' \in PI(P,\varphi(\sigma))$. 
An extension operator is a function $Ext_f$ that takes as input pattern $P$ and an instance pattern $E'$ and returns a new pattern $P'$ according to the extension rule $f$. Specifically, $Ext_f(P, E') = (N \cup V_{f}, \dfr \cup \dfr_{f}, \alpha \cup \alpha^{'}, P)$, where $V_{f}$ is the set of nodes in $\varphi(\sigma)$ satisfying the ordering relation expressed by the extension rule $f$, $\dfr_{f}$ is the set of edges linking the nodes of $V_f$ with the nodes in $E'$, and $\alpha^{'}$ is the labelling function for nodes in $V_f$. 
In this paper, $f \in \{ \dfr, \dfr', \mid \mid, \efr, \efr', dc \}$, which represents respectively:  (1) \emph{direct following}, (2) \emph{direct preceding}, (3)\emph{concurrent}, 4)\emph{eventually following}, (5)\emph{eventually  preceding}, (6) \textit{direct context} relations. 

\begin{sloppypar}
Given $A_{\varphi(\sigma)}$ as adjacency matrix and $R_{\varphi(\sigma)}$ as reachability matrix of size $\mid \sigma \mid-1$ over $E_{\sigma}$, we define {\small $V_{\dfr} = \{ e \in E_{\sigma} \mid \forall n \in E', e \notin E', A_{\varphi(\sigma)}(n, e) = 1 \}$}. In a similar way, {\small $V_{\efr} = \{ e \in E_{\sigma} \mid \forall n \in E', e \notin E', A_{\varphi(\sigma)}(n, e) = 0 , R_{\varphi(\sigma)}(n, e) > 0 \}$} and {\small $V_{\mid \mid} = \{ e \in E_{\sigma} \mid \forall n \in E', e \notin E', A_{\varphi(\sigma)}(n, e) = 0 , R_{\varphi(\sigma)}(n, e) = 0, R_{\varphi(\sigma)}(e, n) = 0 \}$}. Note, $V_{\dfr'}$ and $V_{\efr'}$ can be derived by changing the order of $e$ and $n$ in the definition of $V_{\dfr}$ and $V_{\efr}$, respectively. Finally, $dc$ is defined as $ Ext_{\text{dc}}(P, E') = Ext_{\dfr}(P, E') \cup Ext_{\dfr'}(P, E') \cup Ext_{\mid \mid}(P, E')$. 
\end{sloppypar}

\begin{figure}
    \centering
    \includegraphics[width=\textwidth]{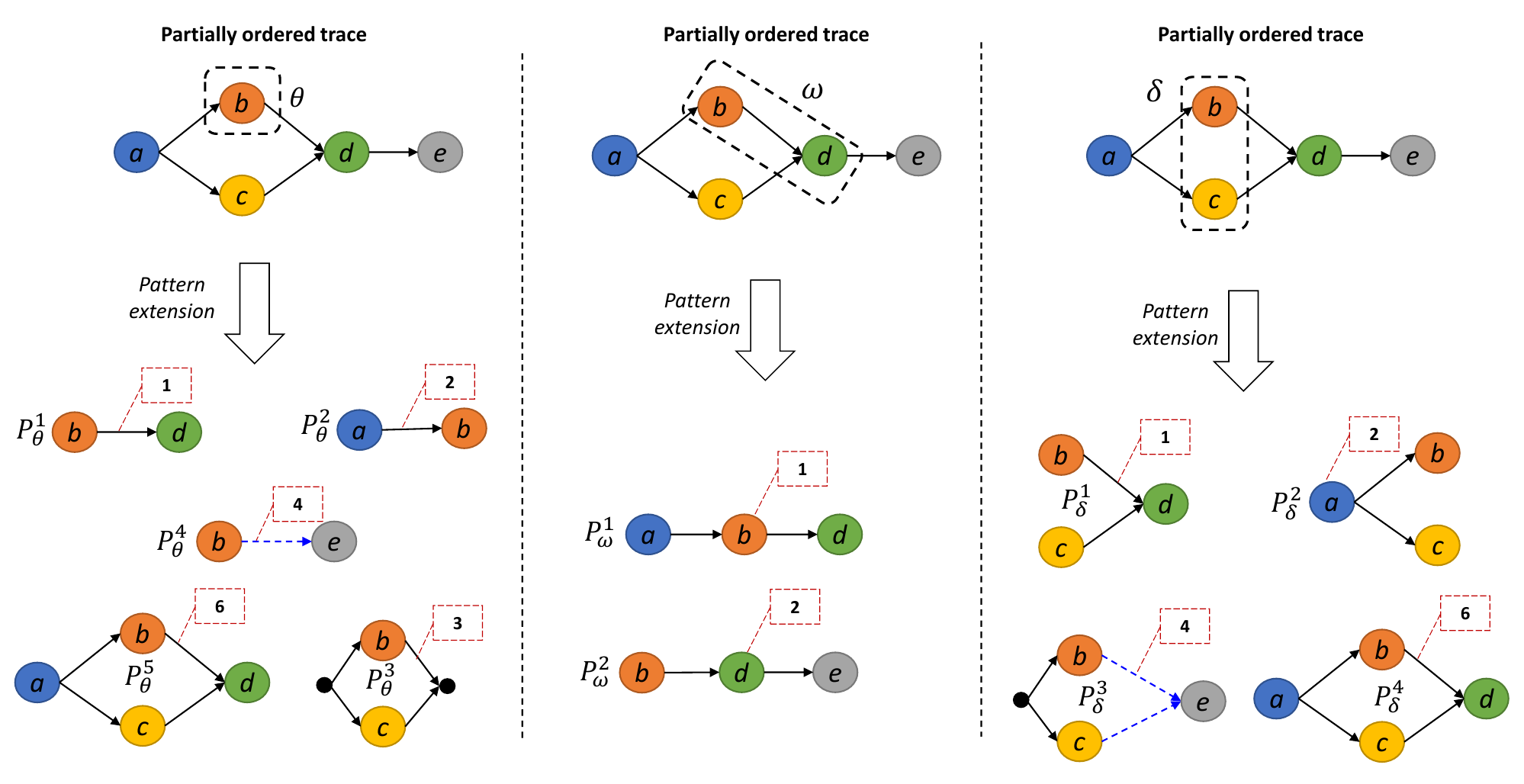}
    \caption{Pattern extension procedure example}
    \label{fig:6}
\end{figure}

\figureautorefname~\ref{fig:6} illustrates some examples of pattern extensions. 
The black dotted boxes in each column of the figure highlight the instance found in the partially ordered trace of a pattern $P$ we want to extend. For instance, single node \textit{b} is an instance of pattern $\theta$, and its corresponding extensions are patterns $P_{\theta}^1, .., P_{\theta}^5$. 
, where the number in the red dotted box reflects the ordering of the rule set. For instance, the first (second) rule represents the \textit{directly following (preceding)} relation, which results in patterns $P_{\theta}^1 ( P_{\theta}^2)$; the third rule involves nodes \textit{concurrent} to b, i.e., in this case, only node \textit{c}\footnote{Please note that we added the black dots only for the sake of clarity in the visualization of a concurrent pattern and they do not belong to the extended pattern.}.; and so on.
Users can select all or a subset of rules $f$ to explore all possibilities for the extension of a selected foundational pattern in each iteration of IMPresseD. Therefore, the set of all extended patterns from the foundational pattern $P$ using a subset of rules called $F'$ is defined as $\mathcal{P}_{P} = \bigcup_{\sigma \in L} \bigcup_{E' \in PI(P,\varphi(\sigma))} \bigcup_{f \in F'} Ext_{f}(P, E')$.

\subsection{Interest functions for outcome-oriented pattern detection}
To show a concrete example of the use of the IMPresseD framework, this section outlines tailored interest functions for the outcome-oriented pattern discovery goal. We designed these functions by analyzing related literature and through discussions with domain experts.

While previous studies in outcome-oriented pattern discovery have focused on identifying patterns that are highly correlated with the outcome~\cite{nguyen2014mining}, we argue that correlation should not be the only dimension of interest. 
Our discussions with healthcare experts 
revealed that 
ignoring the frequency measure may lead to identifying too rare patterns that are often less interesting. 
In addition, 
frequent patterns that are not highly correlated may still be worth exploring. 
For example, a particular treatment \lq \lq A" may be highly frequent but not highly correlated. However, when studying different extensions of \lq \lq A", some interesting correlated patterns may emerge.
Hence, in our analysis, we define frequency-based interest besides correlation-based interest. 
Moreover, it is well-known that potential confounding variables may play an important role in determining the outcome of a treatment process \cite{terada2016significant}. For example, let treatment pattern $P_{1}$ be detected as a pattern that negatively affects the treatment outcome. 
We may find that $P_1$ is only delivered to elderly patients. This questions the reliability of the relation between $P_1$ and the treatment outcome since the patients' age may actually be the real factor leading to worse treatment results.
To mitigate the effect of confounding variables, we consider the distance between cases with or without a specific pattern as the third interest dimension. 

Following these observations, we established three dimensions of interest to support outcome-oriented pattern discovery with their corresponding interest functions. 


\subsubsection{Frequency interest} evaluates the frequency of occurrence of a pattern in the event log. 
In this study, we define \emph{frequency interest function} as the percentage of cases that have at least one pattern instance $P$ as: 

{\emph{CC}$(P, L, \varphi) $}$= \frac{\mid \{ \sigma \in L \mid  PI(P, \varphi(\sigma)) > 0 \}\mid}{\mid L \mid}$

\subsubsection{Outcome interest} measures the effect of each pattern
on the process outcome. For continuous outcome values, we 
use a correlation-based function. For the categorical outcomes, we use an information-gain-based function.  

Let $\Phi$ be a set of values representing possible outcomes. The outcome of a process is defined as a function ${f}: \mathcal{S} \to \Phi$, that maps the set of all possible input traces to the set of all possible outcome values. Then we define $\mathcal{OV} = (f(\sigma))_{\sigma \in L}$ as the outcome vector for event log $L$. Let \emph{PC-freq}$(P, \varphi(\sigma)) = \mid PI(P, \varphi(\sigma)) \mid$ be the frequency of pattern $P$ in trace $\varphi(\sigma)$, we define {\small $\mathcal{FV} = ($\emph{PC-freq}$(P, \varphi(\sigma)))_{\sigma \in L}$} as the frequency vector of pattern $P$ for event log $L$. Then, the \emph{outcome interest function} is defined as {\small $OI(P, L, \varphi) = \rho(\mathcal{OV},\mathcal{FV})$}, where for \emph {\textbf{continues outcome}} $\rho$ is the Spearman correlation coefficient, while for \emph{\textbf{categorical outcome}} $\rho$ is the information gain. 


\subsubsection{Case Distance interest} is designed to mitigate the impact of confounding variables. 
Here, we consider initial case attributes as potential confounding variables. 
Let $\mathcal{AT}$ be a set of user-defined case attributes, $AT_{\sigma_{i}} = (\pi_{d_{j}}(e_{1}))_{d_{j} \in \mathcal{AT}}$ is a vector of initial case attributes corresponding to trace $\sigma_{i}$. Let $C_{P} = \{ \sigma \in L \mid  \mid PI(P, \varphi(\sigma)) \mid > 0 \}$ be the set of cases including an instance of the pattern $P$ and $C_{\bar{P}}= \bigcup_{\sigma \in L} \{ \pi_{case}(\sigma) \} -  C_{P} $ be the set of cases without $P$. Then we define the \emph{Case distance function} as {\small \emph{CD}$(P, L, \mathcal{AT})= \sum_{\sigma_i \in C_{P}} \sum_{\sigma_j \in C_{\bar{P}}} \frac{1}{\mid L \mid} dist(AT_{\sigma_{i}} , AT_{\sigma_{j}})$}. 
Let $dist_{Euc}$ be the \emph{Euclidean} distance for numerical features, and $dist_{Jac}$ be the \emph{Jaccard} distance for $m$ categorical feature, and $F_{normal}$ be a normalization function, then $ dist = \frac{F_{normal}(dist_{Euc}) + dist_{Jac}}{m+1}$ as defined in~\cite{cheung2013categorical}. 

Ideally, there must be \emph{CD}$(P, L, \mathcal{AT})= 0$ to ensure that the pattern $P$ is not influenced by any confounding variable. However, in real-life scenarios, some differences between case attributes are inevitable. To assist users in analyzing which case attributes might have an effect on the outcome, we present a dashboard that visualizes the differences in selected case attributes. This enables the user to pinpoint specific case attributes that may be important for pattern $P$ or explore the reasons behind each process behavior if it is related to the case dimension. An example of this dashboard is presented in \figureautorefname~\ref{fig:5}. 


\section{Implementation and evaluation}\label{IV}

This section aims to demonstrate the usefulness of the IMPresseD framework for a concrete analysis goal defined by expert users (i.e., detecting process patterns affecting the process outcome) through two forms of evaluation. We have implemented an open-source tool in Python for outcome-oriented pattern discovery goals, which is publicly available through GitHub \footnote{https://github.com/MozhganVD/InteractivePatternDetection}. 

The first evaluation (user-based evaluation) aims to show the usefulness of the proposed framework in supporting the user in dealing with pattern discovery in an interactive and multi-interest setting. 
In the second evaluation (quantitative evaluation), we performed a comparative analysis using different sets of patterns in a fully automated setting to evaluate their predictive capabilities. 

\subsection{User based evaluation}
\subsubsection{Evaluation setup.}
The goal of this evaluation is to determine whether our framework is able to discover patterns confirming expert knowledge of the treatment process. 
To this end,
we asked two expert users from the medical domain to use the IMPresseD tool on historical data to discover treatment process patterns affecting patients' survival time. We then asked the users to validate the discovered patterns using their own medical knowledge.

As interest functions, we maximize 
$CC(P, L, \varphi)$ and $OI(P, L, \varphi)$ 
based on the Spearman correlation, and minimize 
$CD(P,L,\mathcal{AT})$.
Regarding 
the visualization dashboard, we opted for \emph{distribution plot} for the numerical features (e.g., age, albumin level, etc.) and \emph{pie chart} for the categorical features (e.g., gender, morphology, etc.) based on expert suggestion. We also visualized the Kaplan-Meier curve, as it is a very common graphical representation of the survival probability for a group of patients based on their observed survival times.
The Log-rank test is also included to check the significance of the difference in survival time between cases with or without a particular pattern.

\subsubsection{Dataset.}
We used an event log provided by the Netherlands Cancer Registry (NCR)  
regarding the treatment process for patients with metastatic stomach or esophageal cancer. 
These patients can usually not be cured and receive palliative care to increase the quality of the remaining lifetime and possibly extend it. Therefore, the outcome of the treatment process is the patient survival time. 

We did some data preprocessing 
according to the domain experts.  
Specially, we removed cases where there were logging errors (e.g., patients for which the survival time was not known), as well as exceptional cases or outliers, like patients who received one or multiple treatment(s) abroad. Similarly, patients with too deteriorated health are removed from the dataset, as they are not fit enough to receive any treatment. At the end of preprocessing, the event logs consisted of 957 cases, 32 distinct treatment codes, and 368 process variants. We also used domain knowledge as a conversion oracle for transforming each trace into a partially ordered trace. In particular, two groups of treatments are considered to be parallel: 1) systematic treatments starting within three days from each other, and 2) all treatments which start and end on the same day. 

\subsubsection{Results.}
\begin{figure}
    \centering
    \includegraphics[width=\textwidth]{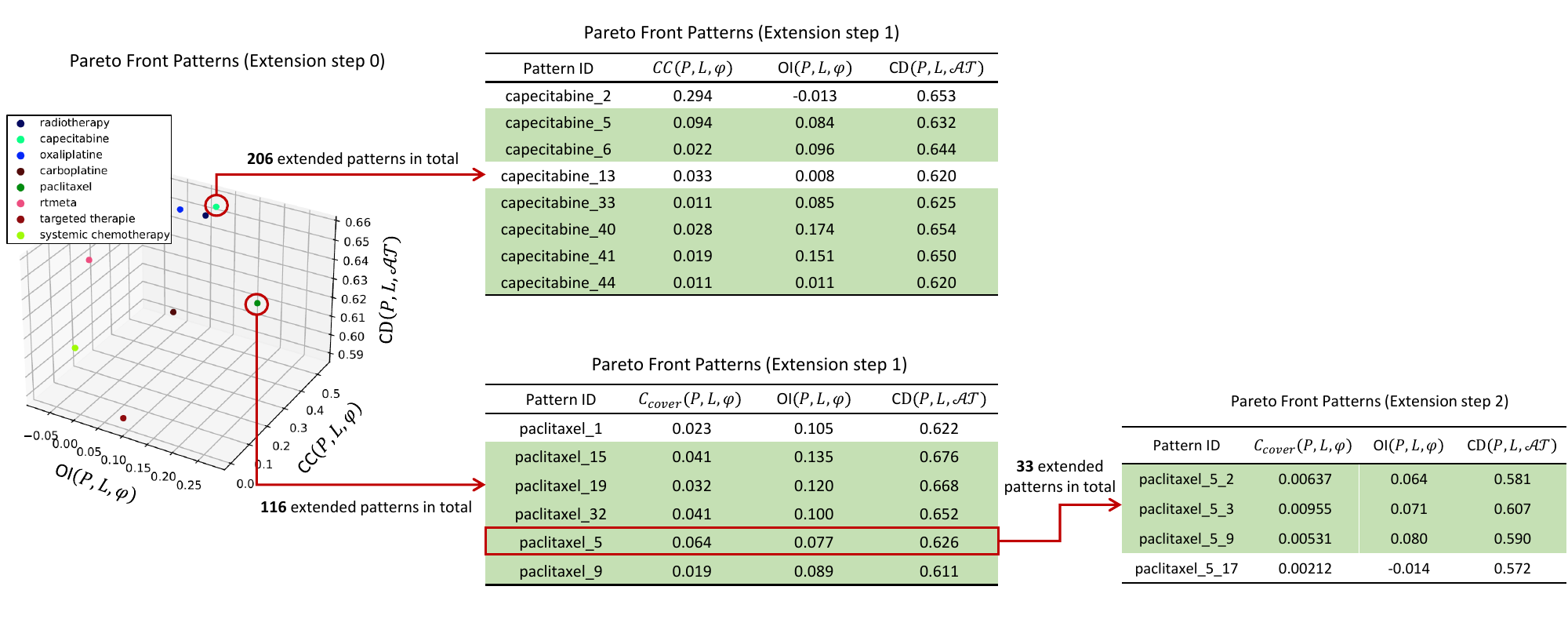}
    \caption{Non-dominated pattern in three iterations of discovery algorithm}
    \label{fig:3}
\end{figure}
In the first iteration of the algorithm (extension step 0), we obtained 8 non-dominated treatments 
as shown in \figureautorefname~\ref{fig:3} (3D graph left side). The framework allows users to assess every single non-dominated treatment in a three-dimensional view.  

\begin{figure}[tb]
    \centering
    \includegraphics[width=\textwidth]{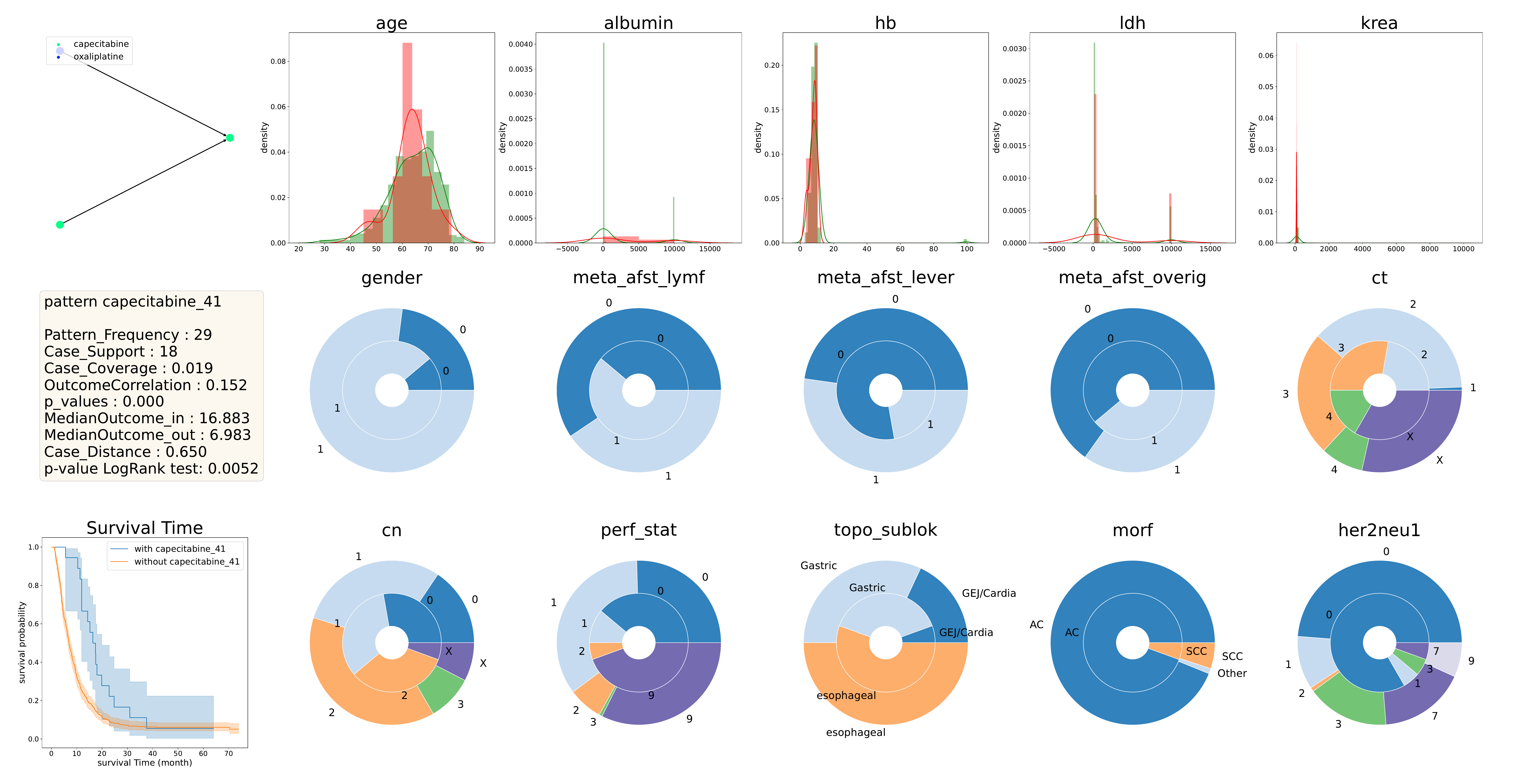}
    \caption{An example of dashboard visualization for a pattern extended from \emph{capecitabine}. Note: the inner ring of pie charts and red color in distribution plots correspond to the cases with the shown pattern in the dashboard.}
    \label{fig:5}
\end{figure}

Users can select each of the recommended single treatments by the Pareto front as a foundational pattern and apply the extension functions discussed in section \ref{III_2}. The experts selected \emph{capecitabine} and \emph{paclitaxel} 
as interesting patterns to extend. 
In the second iteration (extension step 1), we identified 18 patterns out of the 206 extended patterns from \emph{capecitabine} and 14 patterns out of the 116 extended patterns from \emph{paclitaxel} in the Pareto front, indicating that the use of defined interest functions and Pareto front enables users to concentrate on a maximum of 10\% of the total discovered patterns in this step.
The expert decided to filter out patterns with a minimum frequency of 10 patients, thus focusing on the 8 and 6 most frequent patterns from \emph{capecitabine} and \emph{paclitaxel} within the Pareto front, reported in \figureautorefname~\ref{fig:3}. The users decided to stop after one extension step for \emph{capecitabine}, while a second extension step was carried out for \textit{paclitaxel}.
For each pattern, values of each interest function are reported.  Furthermore, we also generate a dashboard showing its control-flow structure and corresponding case data. The main goal of the dashboard is to allow users to compare different case attributes corresponding to the cases with and without patterns. 
These dashboards enable users to investigate the reasons behind each process behavior. An example is shown in \figureautorefname~\ref{fig:5}.
This pattern depicts a treatment pathway that commences with \emph{oxaliplatin} and \emph{capecitabine}. After some time, \emph{oxaliplatin} is stopped, and \emph{capecitabine} is continued. The significant difference between the survival curves of patients with and without this pattern suggests the efficacy of these treatment combinations. Cases with and without patterns are quite similar according to the selected attributes, though the dashboard shows that this pattern was never prescribed to patients with a tumor morphology labeled as \lq \lq other" (which was in line with experts' expectations). 

Patterns colored green in tables inside the \figureautorefname~\ref{fig:3} are the patterns marked as interesting by expert users (i.e., patterns validated by medical knowledge). The users considered two patterns from extending \emph{capecitabine} not interesting because of a too low correlation with the outcome, leading to very similar survival time for patients with and without the pattern (MedianOutcome\textunderscore in/out in the dashboard), which does not allow them to say anything about the relationship with the outcome. 
As regards patterns extended from \textit{paclitaxel},
in the first step only one pattern was marked as not interesting. The reason is that the user expected an additional treatment which, however, was not possible to detect in combination with the discovered patterns. Further investigations are needed to determine why the occurrence of this particular treatment in the dataset does not fit with experts' expectations. Note that for the second extension, with foundational pattern \textit{paclitaxel\_5}, the users were especially interested in extensions involving radiotherapy. The last extended pattern in the extension step 2 did not involve radiotherapy and was hence marked as not interesting.
Overall, the detected patterns confirmed the effectiveness of the previously known combination of treatments, providing valuable evidence-based insight. Only a few patterns were marked as not interesting. Both users found the visualization dashboard very helpful in understanding the detected patterns and in uncovering potential relations with the case attributes.
We would like to point out that without using the Pareto front, users would have to either try different thresholds or explore all the extended patterns manually. 

\subsection{Quantitative evaluation}
\subsubsection{Evaluation setup.} 
The goal of the quantitative evaluation
consists in assessing the predictive capabilities of patterns detected employing multi-interest functions
compared to patterns detected utilizing a single dimension or without any filtering. If the multi-interest functions obtain a predictive performance in line with the other strategies, this shows they allow to preserve the same predictive power, in addition to them leading to more meaningful process patterns filtering out many non-interesting ones (as illustrated in the user-based evaluation).
We drew inspiration from the common evaluation used in\lq\lq deviance mining"~\cite{nguyen2014mining} to assess the quality of the set of discovered patterns in predicting the outcome of the process without exploiting the user's knowledge. 
In this setting, discovered patterns are treated as independent features, while the process's outcome is considered the dependent feature. Frequency-based encoding is used to encode independent features. 
Specifically, we compare the performance of decision trees (DTs) trained on the $K$ patterns obtained from the Pareto front
in each extension step to those trained on the \emph{top} $K$ patterns identified by considering every single dimension, as well as those trained on all discovered patterns. To achieve this, 
all the $K$ non-dominated patterns in the extension step $i^{th}$ were used as foundational patterns to be extended in iteration $i+1^{th}$. As interest functions, we maximize the outcome (information-gain-based) and frequency functions and minimize the case distance function.
During the pattern discovery procedure, we only considered the training set 
to prevent potential bias or information leakage in the evaluation.


\subsubsection{Datasets.}
We analyzed the three most commonly used event logs in outcome prediction literature, namely \textit{BPIC2012}, \textit{BPIC2011}, and \textit{Production}, by leveraging preprocessed and labeled logs from prior research
\cite{teinemaa2019outcome}. We used all case-related attributes for calculating the case distance function.
For the NCR dataset, we divided the survival time into three classes with equal frequency based on the experts' knowledge for this evaluation.

\begin{figure}[tb]
    \centering
    \includegraphics[width=\textwidth]{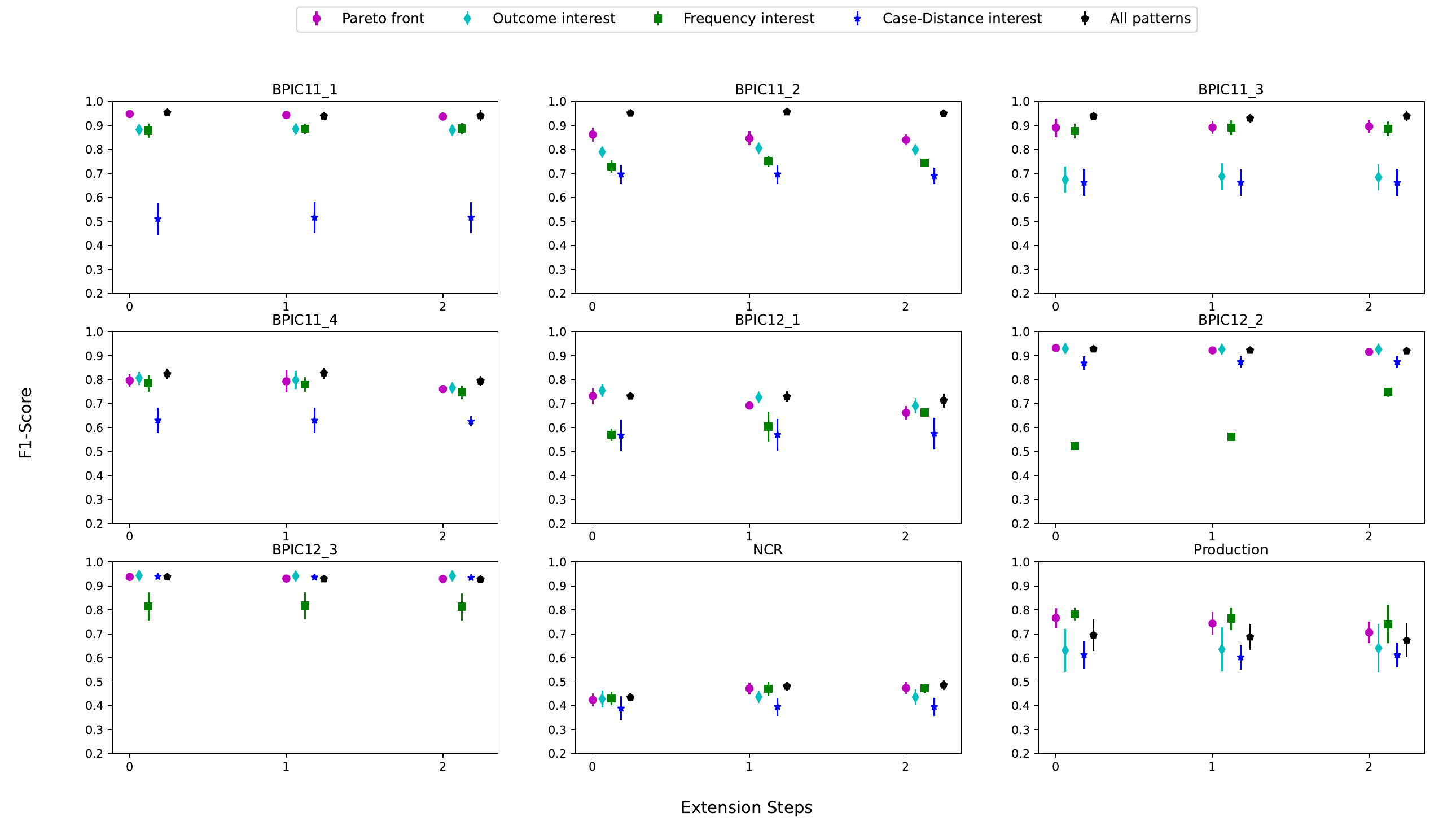}
    \caption{Quantitative evaluation results}
    \label{fig:2}
\end{figure}

\subsubsection{Results.}

\figureautorefname~\ref{fig:2} presents the results of the 5-fold cross-validation (i.e., the average F1-score with minimum and maximum obtained values). 
The DT trained on the patterns obtained from the Pareto front outperformed or it is as accurate as its counterparts. The only configuration that does better in some cases is the all\_patterns configuration, which involves a much higher number of patterns. Indeed, on average, the ratio between the size of the feature set obtained from the Pareto front and the size of the feature set obtained in all\_patterns configuration is 47.5\%. This result shows that using Pareto optimal solutions, we combine the best of multiple criteria and manage to retain discriminative information with a smaller number of patterns than all possible ones. Another interesting finding is that the results of the DT trained on patterns from the Pareto front consistently rank among the best ones, while results of DTs trained on patterns obtained from single dimensions show a stronger dependency on the dataset.

When comparing single-interest measures, the case distance obtained the worst performance in most of the tested datasets. The \emph{outcome measure} outperformed all single measures in 5 out of 10 studied event logs (BPIC11\textunderscore2, BPIC11\textunderscore4, BPIC12\textunderscore1, BPIC12\textunderscore2, BPIC12\textunderscore3), while the \emph{frequency interest} outperformed the other single measures 
 in 3 event logs (BPIC11\textunderscore3, NCR, Production). This suggests that there might be a relationship between the characteristics of the event log and the predictive power of single-interest dimensions.

\subsection{Discussion}
The quantitative evaluation indicates that using the Pareto front 
leads to comparable or better prediction performance than the ones achieved by using single measures, and with much fewer patterns than using all possible ones. Using the Pareto front also has the additional advantage that less effort is required 
than selecting a threshold for a specific metric. Furthermore, the developed approach provides a flexible means for the user to define the desired pattern characteristics.
Note that the quantitative evaluation also shows that the proposed method has the potential to be used in a fully automated setting.

However, a surprising observation is that extending the process patterns often does not improve the prediction results, except for a slight improvement in performance after the first extension in the NCR dataset. This may be due to an overlap between the pattern obtained from the $(i+1)^{th}$ iteration and the foundational patterns in the $i^{th}$ iteration. 
Considering all patterns in the Pareto front as foundational patterns for being extended in the next iteration may have led to overlap that increases the dimension of the problem without adding much new information. One direction for future research would be to minimize the overlap between patterns obtained from each iteration.

On the other hand, the results of the user-based evaluation demonstrate the usefulness of the IMPresseD framework in discovering process patterns for 
supporting outcome-oriented process pattern detection. The Pareto front selection of patterns allows users to reach their desired pattern without exploring many non-interesting patterns. Furthermore, the designed visualization dashboard provides effective support to the human analyst in exploring and interpreting the patterns. We would like to point out that, to the best of our knowledge, no other process pattern discovery tool provides these functionalities. However, this evaluation has some threats to validity. First, being based on a use case, these results cannot be generalized to different contexts. Furthermore, only two experts were involved in verifying the discovered patterns. To mitigate these threats, we provide a prototype of the tool to enable other researchers to replicate our results and apply the approach to other case studies. Furthermore, a comprehensive survey involving more experts from different perspectives, such as data scientists and oncologists, is planned to evaluate the proposed method on a wider scale. 
 
\section{Conclusion and future work}\label{V}
The paper presented the IMPressed framework, designed to derive interesting and easily interpretable process patterns for the end users. 
The framework is iterative and interactive and allows the user to select the most interesting patterns to expand further. The paper also discussed a concrete analysis goal of deriving process patterns affecting the process outcome, which is a complex problem that requires considering different aspects. The paper evaluated the proposed approach using a real-life case study in healthcare and in a completely automated setting using publicly available event logs. 
Overall, the paper contributes to the process pattern discovery literature by introducing a framework that takes into account a multi-dimensional notion of interest and by demonstrating its effectiveness through empirical evaluations.
In future work, to further evaluate and enhance the efficacy of our proposed framework, we intend to conduct a comprehensive survey that draws on a wider range of expert knowledge and opinions. This survey will allow us to gather valuable feedback on the usefulness of our framework and explore potential avenues for future research. Additionally, we intend to explore additional extension operators to discover more complex patterns, as well as introduce constraints on the pattern extension in a fully-automated setting.

\bibliographystyle{splncs04}
\bibliography{samplepaper}

\end{document}